%% file: main.tex
\documentclass[conference]{IEEEtran}
\IEEEoverridecommandlockouts
\usepackage{cite}
\usepackage{amsmath,amssymb,amsfonts,amsthm}
\usepackage{dsfont}
\usepackage{graphicx}
\usepackage{textcomp}
\usepackage{xcolor}
\usepackage{hyperref}
\usepackage{eucal}
\usepackage{bm}

\usepackage{caption}
\usepackage{subcaption}

\newtheorem{proposition}{Proposition}
\newcommand{\arl}[1]{\text{ARL}_{#1}}
\newcommand{\boldarl}[1]{\text{\textbf{ARL}}_{\mb{#1}}}

\newcommand{\skl}{\mathit{sKL}}

\usepackage{hhline}
\usepackage{algorithm}
\usepackage{algpseudocode}

\newcommand{\MyComment}[1]{\State \textcolor{gray}{// #1}}
\newcommand{\mb}[1]{\mathbf{#1}}

\usepackage{pgf,tikz,pgfplots,pgfplotstable}
\pgfplotsset{compat=1.15}
\usepgfplotslibrary{statistics}
\usepgfplotslibrary{groupplots}

\def\BibTeX{{\rm B\kern-.05em{\sc i\kern-.025em b}\kern-.08em
    T\kern-.1667em\lower.7ex\hbox{E}\kern-.125emX}}
\begin{document}

\input{figures/setup_tikz}

\title{Class Distribution Monitoring for Concept Drift Detection
}

% \author{\IEEEauthorblockN{Anonymous Author(s)}
% \IEEEauthorblockA{\textit{Anonymous Institute}\\
% City, Country\\
% e-mail address
% }
%}
% \and
\author{
\IEEEauthorblockN{Diego Stucchi}
\IEEEauthorblockA{%\textit{Dipartimento di Elettronica, Informazione e Bioingegneria} \\
\textit{Politecnico di Milano}\\
Milan, Italy\\
diego.stucchi@polimi.it}
\and
\IEEEauthorblockN{Luca Frittoli}
\IEEEauthorblockA{%\textit{Dipartimento di Elettronica, Informazione e Bioingegneria} \\
\textit{Politecnico di Milano}\\
Milan, Italy\\
luca.frittoli@polimi.it}
\and
\IEEEauthorblockN{Giacomo Boracchi}
\IEEEauthorblockA{%\textit{Dipartimento di Elettronica, Informazione e Bioingegneria} \\
\textit{Politecnico di Milano}\\
Milan, Italy\\
giacomo.boracchi@polimi.it}
}
% \and
% \and
% \and

\maketitle

\begin{abstract}
% We introduce Class Distribution Monitoring (CDM), an effective concept-drift detection scheme that monitors a datastream by class-specific change-detection tests. Our solution leverages multiple instances of an online and nonparametric change-detection algorithm based on QuantTree to monitor the class-conditional distributions of supervised samples.
We introduce Class Distribution Monitoring (CDM), an effective concept-drift detection scheme that monitors the class-conditional distributions of a datastream. In particular, our solution leverages multiple instances of an online and nonparametric change-detection algorithm based on QuantTree.
CDM reports a concept drift after detecting a distribution change in any class, thus identifying which classes are affected by the concept drift. This can be precious information for diagnostics and adaptation. Our experiments on synthetic and real-world datastreams show that when the concept drift affects a few classes, CDM outperforms algorithms monitoring the overall data distribution, while achieving similar detection delays when the drift affects all the classes. Moreover, CDM outperforms comparable approaches that monitor the classification error, particularly when the change is not very apparent. Finally, we demonstrate that CDM inherits the properties of the underlying change detector, yielding an effective control over the expected time before a false alarm, or Average Run Length ($\arl{0}$).
\end{abstract}


\section{Introduction}
% What is the problem?
Datastreams represent a challenging scenario for machine learning models~\cite{bahri2021data} since their distribution might change over time, resulting in a \emph{concept drift}~\cite{lu2018learning}. This phenomenon has been widely studied in settings where the drift worsens the performance of a classifier, which must be adapted to the new data distribution. 
% Why is it hard? (E.g., why do naive approaches fail?) / Why is it interesting and important?
To this purpose, most solutions monitor the classification error, ignoring drifts that have little impact on the error rate, which are called \emph{virtual drifts}. However, in practical situations such as in industrial monitoring, any distribution change in streaming data should be promptly detected for diagnostic purposes. Moreover, in the emerging field of \emph{open-set recognition}~\cite{geng2020recent}, a classifier is required to recognize the occurrence of known classes and also to detect samples that do not belong to any known class, thus it is crucial to update the decision boundary of the classifier even when the accuracy on known classes does not decrease. This enables updating also the regions in which the classifier predicts with low confidence, where unknown samples might appear. % \di{In particular, this enables updating the boundary of regions where the classifier has low confidence, and where unknown samples might appear.}

% Why hasn't it been solved before? (Or, what's wrong with previous proposed solutions? How does mine differ?)
Most concept drifts can be detected by monitoring the data distribution by \emph{online change-detection tests}~\cite{basseville1993}, which is another common approach in concept-drift detection~\cite{lu2018learning}. However, none of these methods can exploit supervised information since they overlook class labels. Our intuition is that class labels can be included in statistically sound change-detection tests to monitor the class-conditional distributions instead of the overall data distribution. To the best of our knowledge, this approach has never been investigated before.
% Moreover, drifts that affect only a subset of classes can be hard to detect since the rest of the data still follows the initial distribution.

% What are the key components of my approach and results?
We fill this gap by proposing \emph{Class Distribution Monitoring} (CDM)\footnote{Our code is available at \textcolor{blue}{\url{https://boracchi.faculty.polimi.it/Projects}}. This paper is part of the Proceedings of the International Joint Conference on Neural Networks \copyright 2022 IEEE, DOI: 10.1109/IJCNN55064.2022.9892772.}, in which we employ separate instances of \emph{QuantTree Exponentially Weighted Moving Average} (QT-EWMA)~\cite{frittoli2021change} to monitor the class-conditional distributions. QT-EWMA is a nonparametric online change-detection test based on QuantTree histograms~\cite{boracchi2018quanttree}, and is designed to monitor multivariate datastreams. We report a concept drift after detecting a change in the class-conditional distribution of at least one class. The main advantages of CDM are: \emph{i)} it can detect any relevant drift, including virtual ones that have little impact on the classification error and are by design ignored by methods that monitor the error rate of a classifier; \emph{ii)} it can detect concept drifts affecting only a subset of classes more promptly than methods that monitor the overall data distribution, since the other class-conditional distributions do not change; \emph{iii)} it provides insights on which classes have been affected by concept drift, which might be crucial for diagnostics and adaptation; \emph{iv)} it effectively controls false alarms by maintaining a target \emph{Average Run Length} ($\arl{0}$), i.e., the expected time before a false alarm~\cite{basseville1993}, which can be set before monitoring. %Unfortunately, the vast majority of concept drift detection methods fail to control the $\arl{0}$ effectively.

To summarize, our main contributions are:
\begin{itemize}
    \item We introduce Class Distribution Monitoring (CDM), a novel online and nonparametric monitoring scheme for concept-drift detection leveraging supervised samples.
    \item Our CDM can, by design, detect drifts affecting only a subset of classes and, contrarily to most concept drift detectors, identify the drifted classes. %which classes have been affected by concept drift. %, which can be very useful for diagnostic.
    \item We theoretically and empirically demonstrate that CDM can be configured to yield the desired $\arl{0}$, thus effectively controlling false alarms, even though it employs several change-detection tests simultaneously.
\end{itemize}
Our experiments on synthetic and real-world datastreams show that CDM outperforms algorithms monitoring the overall distribution when the concept drift affects only a subset of classes, while achieving comparable detection delays when the change affects all classes. CDM can also effectively detect virtual drifts, which are ignored by methods that monitor the classification error but might be relevant in practice.

\section{Problem Formulation}
% Let us consider a datastream $x_1,x_2,\ldots\in\mathbb{R}^d$ in which each sample $x_t$ is paired with a class label ${\mc{L}(x_t)=y_t\in\{1,\ldots,M\}}$.
We address the problem of detecting a concept drift in a virtually unlimited datastream $\{(x_t, y_t)\}$, where each sample $x_t\in\mathbb{R}^d$ is associated to a class label $y_t\in\{1,\ldots,M\}$. We assume that the observations $x_t$ are independent realizations of a random vector that follows an initial distribution $\phi_0$. We denote by $\phi_0^m$ the \emph{class-conditional} distribution, i.e., the distribution of instances belonging to class $m$, defined by 
\begin{equation}
    \mathbb{P}_{\phi_0^m}(x_t) = \mathbb{P}_{\phi_0}(x_t \mid y_t=m),\label{eq:class_prob}
\end{equation}
for each $m\in\{1,\ldots,M\}$. In other words, we say that $x_t \sim \phi_0$ if and only if $x_t \sim \phi_0^m$, where $y_t=m$. We assume that a concept drift affects at least one class-conditional distribution, resulting in a change $\phi_0^m \to \phi_1^m$ occurring at an unknown time $\tau$ for some $m\in\{1,\ldots,M\}$. We assume that an annotated dataset $TR$ sampled from the initial distribution $\phi_0$ is provided before monitoring, to configure the concept-drift detector. 

In the concept-drift detection literature \cite{gama2004learning,frias2014online,de2018wilcoxon,ross2011nonparametric} it is usually assumed that, during monitoring, the true labels $y_t$ are revealed after the prediction made by a classifier $\mathcal{K}$, to provide immediate feedback on whether the classification was correct. We operate in the same settings, even though in practical situations the labels are typically provided only for a few samples of the datastream. In the latter case, methods that require the true labels can take as input only those samples $x_t$ for which the label $y_t$ is provided.

The goal of a concept-drift detection algorithm is detecting any distribution change as soon as possible by analyzing the incoming samples. We indicate by $t^*$ the detection time, and we measure the detection performance by the detection delay $t^*-\tau$. A crucial challenge in change detection is controlling false alarms, which in online settings means maintaining a target \emph{Average Run Length} ($\arl{0}$), defined as 
\begin{equation}
    \arl{0}=\mathbb{E}_{\phi_0}[t^*],\label{eq:arl_def}
\end{equation}
which is the expected time before having a false alarm, namely a detection % when there are no changes
that does not correspond to any distribution change~\cite{basseville1993}. The $\arl{0}$ represents the online counterpart of the false positive rate in statistical hypothesis testing. Operating at a controlled $\arl{0}$ allows to limit the frequency of false alarms, which typically trigger costly adaptation procedures such as re-training a classifier. This is particularly important in industrial monitoring, and in experimental test-bed to enable a fair comparison between different concept-drift detectors. Unfortunately, the vast majority of concept-drift detection methods fail to control the $\arl{0}$ effectively.

% on a subset of classes $\mc{C}$, namely:
% \begin{equation}
% x_t\sim \phi_1^{y_t}
% \begin{cases}
%      \neq \phi_0^{y_t}, & \text{if } y_t \in \mc{C} \\
%      = \phi_0^{y_t}, & \text{otherwise}
% \end{cases}
% ,
% \end{equation}
% where $\mc{C}$ must contain at least a class label, and $t\geq\tau$. %\di{Si capisce in cosa consiste tau?}
% A concept drift is defined as a class conditional distribution change $\phi_0^m\rightarrow\phi_1^m$ occurring at an unknown time $\tau$ for some $m\in\{1,\ldots,M\}$ (at least one). 

% The goal is to detect a concept drift as soon as possible by analyzing the datastream one sample at a time while controlling false alarms by setting the Average Run Length ($\arl{0}$), i.e., the expected time before a false alarm.

\section{Related Work}
Concept-drift detection \cite{lu2018learning} is a challenging problem in datastream learning, and has been addressed in different settings and by different approaches, which we summarize here. %Following the vast majority of the literature, we consider supervised settings where the labels $y_t$ of of the data samples $x_t$ are assumed to be known. 
Since we focus only on concept-drift detection, we do not review the literature on \emph{concept-drift adaptation}. We refer to~\cite{gama2014survey} for a survey on this subject.

The most popular concept-drift detection methods analyze the binary stream $\{e_t\}$ defined by the errors of a classifier $\mathcal{K}$:
\begin{equation}
    e_t=\mathds{1}(\mathcal{K}(x_t)\neq y_t),\label{eq:errors}
\end{equation}
% and report a concept drift when the error rate increases. In particular, \emph{Drift Detection Method} (DDM)~\cite{gama2004learning} and its variants \emph{Early Drift Detection Method} (EDDM)~\cite{baena2006early} and \emph{Hoeffding's inequality-based Drift Detection Method} (HDDM)~\cite{frias2014online} employ statistical tests to assess whether the error rate has changed significantly. \emph{EWMA for Concept Drift Detection} (ECDD)~\cite{ross2012exponentially} analyzes $\{e_t\}$ online by an \emph{Exponentially Weighted Moving Average} (EWMA) chart~\cite{roberts1959control}. Contrarily to the vast majority of concept drift detection methods, ECDD allows to control false alarms by setting the $\arl{0}$. Thanks to this property, in our experiments we can fairly compare ECDD and our solution by configuring them to maintain the same $\arl{0}$.
and report a concept drift when the error rate increases. In particular, \emph{Drift Detection Method} (DDM)~\cite{gama2004learning} and its variants~\cite{frias2014online,de2018wilcoxon} apply statistical tests on recent windows of $\{e_t\}$ to assess whether the error rate has increased significantly. Moreover, it has recently been proposed to monitor the classification performance on individual classes to handle imbalanced datastreams and drifts affecting only some classes~\cite{wang2020auc,korycki2021concept}. However, none of these solutions can be configured to maintain the $\arl{0}$. In contrast, \emph{EWMA for Concept Drift Detection} (ECDD)~\cite{ross2012exponentially} analyzes $\{e_t\}$ online by an \emph{Exponentially Weighted Moving Average} (EWMA) chart~\cite{roberts1959control}, which enables controlling false alarms by setting the $\arl{0}$. Thanks to this property, in our experiments we can fairly compare ECDD and our solution by configuring them to maintain the same $\arl{0}$.

Another relevant class of concept drift detection methods monitors the distribution of the input data, overlooking the information possibly coming from class labels, and therefore can operate also when few or no labels are available. These methods leverage online change-detection tests~\cite{basseville1993} to analyze the data distribution over time. A popular approach consists in monitoring the likelihood of the streaming data with respect to a density model such as a Gaussian~\cite{guralnik1999event} or a Gaussian Mixture~\cite{kuncheva2011change,alippi2015change}. The main limitation of these approaches is the assumption that $\phi_0$ can be approximated by a distribution from a known family, which might not be the case when dealing with real-world data. %Moreover, among these solutions, the only one in which the $\arl{0}$ can be configured before monitoring is SPLL-CPM~\cite{frittoli2021change}, where the log-likelihood of the Gaussian Mixture is monitored by a nonparametric Change Point Model (CPM) controlling the $\arl{0}$~\cite{ross2011nonparametric}.

A very flexible nonparametric approach consists in modelling the initial data distribution by a histogram~\cite{boracchi2018quanttree,frittoli2021change,lau2018binning}, and then monitoring the proportion of incoming samples that falls in each bin of the histogram. For instance, QuantTree~\cite{boracchi2018quanttree} computes the Pearson test statistic~\cite{lehmann2006testing} over fixed-size batches, while its extension QT-EWMA~\cite{frittoli2021change} enables online monitoring controlling the $\arl{0}$. Other nonparametric online change detectors are either based on PCA~\cite{kuncheva2013pca,qahtan2015pca}, permutation tests~\cite{ho2005martingale,mozafari2011precise}, or the Maximum Mean Discrepancy statistic (MMD)~\cite{gretton2012kernel} computed over sliding windows~\cite{li2015m,keriven2020newma}. However, among these, the only one that can control the $\arl{0}$ regardless of the initial data distribution is Scan-B~\cite{li2015m}.

%\lu{vogliamo citare anche multiple hypothesis testing (JIT \cite{alippi2013just}, altro?). Nella survey di riferimento~\cite{lu2018learning} ne parlano come di una terza categoria di concept drift detectors.}

\section{Proposed Solution}
Here we briefly introduce QT-EWMA~\cite{frittoli2021change} (Section~\ref{subsec:detectors}), which is the change-detection algorithm we use to define our solution. Then, we present Class Distribution Monitoring (CDM) (Section~\ref{subsec:cdm}). Finally, we demonstrate that CDM inherits the properties of QT-EWMA and analyze its computational complexity (Section~\ref{subsec:arl}). In particular, we show that CDM can control false alarms by yielding the desired $\arl{0}$.

\subsection{Concept Drift Detection by Distribution Monitoring}\label{subsec:detectors}
Most concept-drift detection methods that monitor the distribution of the datastream $\{x_t\}$ compute at each time $t$ a test statistic $T_t$, and report a drift after detecting a distribution change~\cite{lu2018learning}. Typically, a change is detected when $T_t>h_t$, where $h_t$ is a threshold defined to control the probability of having a false alarm. The detection time $t^*$ is defined as the first time $t$ in which the statistic exceeds the threshold. %In online monitoring, false alarms should be controlled by setting the Average Run Length ($\arl{0}$), which is defined as $\arl{0}=\mathbb{E}_{\phi_0}[t^*]$, namely the expected time before a false alarm~\cite{basseville1993}. %However, there are not many online change detectors that can effectively control the $\arl{0}$, especially among those monitoring multivariate datastreams. Among these, 
We adopt QuantTree Exponentially Weighted Moving Average (QT-EWMA)~\cite{frittoli2021change}, which effectively controls the $\arl{0}$ and is also completely nonparametric, i.e., it does not require any assumption on the initial data distribution $\phi_0$.

QT-EWMA models $\phi_0$ by a QuantTree histogram~\cite{boracchi2018quanttree} built on the training set $TR$. The histogram is defined by ${Q=\{(S_k,\pi_k)\}_{k=1}^K}$, where $S_k$ are the histogram bins, $\pi_k$ the corresponding target bin probabilities, and $K$ is the number of bins to be set a priori. Then, QT-EWMA monitors the proportion of samples falling in each bin of the histogram by $K$ EWMA statistics~\cite{roberts1959control}:
\begin{equation}
    Z_{k,t} = (1-\lambda) Z_{k,t-1} + \lambda b_{k,t},\quad Z_{k,0} = \pi_k,\label{eq:ewmas}
\end{equation}
%where $b_{k,t} = \mathds{1}(x_t\in S_k)$ indicates
where the binary statistics $b_{k,t} = \mathds{1}(x_t\in S_k)$ indicate the bin of the histogram in which $x_t$ falls, for $k\in\{1,\ldots,K\}$. Then, the statistic $T_t$ is defined by
\begin{equation}
    T_t = \sum_{k=1}^{K} \dfrac{(Z_{k,t}-\pi_k)^2}{\pi_k}.\label{eq:qtewma}
\end{equation}
Each statistic $Z_{k,t}$ is an incremental measure of the proportion of samples acquired until time $t$ falling in each bin $S_k$. The statistic $T_t$ assesses how much the $Z_{k,t}$ deviate from the target bin probabilities $\pi_k$, thus it is similar to the Pearson statistic~\cite{lehmann2006testing}. %Full implementation details can be found in~\cite{frittoli2021change}. 
The main advantage of this solution is that the distribution of $T_t$ \eqref{eq:qtewma}, like any other statistic based exclusively on the number of points falling in the bins of a QuantTree histogram, is independent from $\phi_0$, as demonstrated in~\cite{boracchi2018quanttree}. This property enables nonparametric monitoring, and allows to define thresholds $\{h_t\}$ for QT-EWMA such that:
\begin{equation}
    \mathbb{P}_{\phi_0}(T_t>h_t \mid T_k\leq h_k \forall k<t) = \alpha,
    \label{eq:alpha}
\end{equation}
which have been shown to guarantee a desired $\arl{0}$ when $\alpha=1/\arl{0}$~\cite{margavio1995alarm}. These thresholds are computed by Monte Carlo simulations that are described in detail in~\cite{frittoli2021change}.

\subsection{Class Distribution Monitoring}\label{subsec:cdm}
%\lu{dobbiamo decidere se limitarci a QT-EWMA o parlare di class-specific monitoring in modo che ci rientri anche SPLL-CPM}
QT-EWMA, like other concept-drift detectors that monitor the data distribution, is designed to operate in unsupervised settings, and therefore ignores the labels $y_t$, which we assume to be regularly provided during monitoring. As a result, concept drifts affecting only a subset of classes can be hard to detect following this approach.

To exploit class labels, we propose Class Distribution Monitoring (CDM), which is illustrated in Algorithm~\ref{alg:cdm}. First, we divide the training set $TR$ into $M$ subsets $TR^m$ and use these to construct $M$ QuantTree histograms $Q^m=\{(S^m_k,\pi_k)\}$~\cite{boracchi2018quanttree}, corresponding to the classes $m\in\{1,\ldots,M\}$ (lines \ref{line:TRm}--\ref{line:QTm}). When an input sample $x_t$ is provided with its label $y_t$, we find the histogram bin such that $x_t\in S^m_k$ in the QuantTree $Q^m$ corresponding to its label $m=y_t$ (line \ref{line:yj}). Then, we compute the QT-EWMA statistic $T^m_{t_m}$ \eqref{eq:qtewma} (lines \ref{line:ewma}--\ref{line:qt_ewma}), where $t_m$ is the number of samples of class $m$ observed until time $t$. We report a concept drift as the first time $t$ when $T^m_{t_m}>h_{t_m}$, where $h_{t_m}$ is the QT-EWMA threshold defined by \eqref{eq:alpha} (lines \ref{line:start_det}--\ref{line:end_check}). We remark that, contrarily to the other concept-drift detectors, our algorithm returns, on top of the detection time $t^*$, the class $m^*$ that triggered the detection (line \ref{line:end_det}).

\begin{algorithm}[t]
\centering
\caption{Class Distribution Monitoring (CDM)}
\label{alg:cdm}
\begin{algorithmic}[1]
\Require datastream $\{(x_t,y_t)\}_t$, target probabilities $\{\pi_k\}_{k=1}^{K}$, thresholds $\{h_t\}_t$, $TR=\{(x,y)\}$
\Ensure detection flag $\texttt{ChangeDetected}$, detection time $t^*$, drifted class $m^*$
\MyComment{Configuration:}
\State $\texttt{ChangeDetected}\gets \text{False}, t^* \gets \infty, m^*\gets 0$\;
\For{$m=1,\ldots,M$}
    \State $TR^m \gets \{x: (x,y)\in TR,y=m\}$\label{line:TRm}
    \State build QuantTree $Q^m=\{(S^m_k,\pi_k)\}$~\cite{boracchi2018quanttree} from $TR^m$\label{line:QTm}
    % \State $t_m\gets0$
    \State initialize $t_m\gets0$, $Z^m_{k,0}\gets \pi_k, k\in\{1,\ldots,K\}$
\EndFor
\MyComment{Monitoring:}
\For{$t=1,\ldots$} 
    \If{the label $y_t$ is provided}
        \State $m \gets y_t,\; t_m\gets t_m+1,\; b_{k,t_m}\gets \mathds{1}(x_t\in S^m_k)$\label{line:yj}
        %\State $\mathcal{X}^m\gets [\mathcal{X}^m, x_t],\; t_m\gets t_m+1$\label{line:substream}
        \State compute $Z^m_{k,t_m}$ \eqref{eq:ewmas} for $k\in\{1,\ldots,K\}$\label{line:ewma}
        \State compute QT-EWMA statistic $T^m_{t_m}$ \eqref{eq:qtewma}\label{line:qt_ewma}
        \If{$T^m_{t_m} > h_{t_m}$}\label{line:start_det}
            \State $\texttt{ChangeDetected}\gets \text{True}$
            \State $t^* \gets t,\; m^*\gets m$
            \State \textbf{break}
        \EndIf\label{line:end_check}
    \EndIf
\EndFor
\State \textbf{return} $\texttt{ChangeDetected}, t^*, m^*$\label{line:end_det}
\end{algorithmic}
\end{algorithm}

\subsection{Properties of CDM}\label{subsec:arl}
Here we illustrate the most important properties of CDM, in particular the control of the $\arl{0}$, and analyze its computational complexity. 

\textbf{Online and Nonparametric Monitoring.} Consistently with the notation introduced in Section~\ref{subsec:detectors}, we can see CDM as an online change-detection test with statistic $\Tilde{T}_t$ defined as
\begin{equation}
    \tilde T_t=T^m_{t_m},\quad m=y_t,\label{eq:CDMstat}
\end{equation}
and thresholds $\tilde h_t=h_{t_m}$. %CDM inherits three fundamental properties from QT-EWMA: it is nonparametric, it operates online, and controls false alarms by maintaining the target $\arl{0}$ set before monitoring. 
The online nature of CDM is evident from Algorithm~\ref{alg:cdm}, where the datastream is processed one sample $(x_t,y_t)$ at a time. The nonparametric nature of CDM derives from the fact that the distribution of the test statistic $\tilde T_t$, like any other statistic based on QuantTree, does not depend on the initial distribution $\phi_0$, as shown in~\cite{boracchi2018quanttree}.

% In online change detection false alarms are typically controlled by setting the Average Run Length ($\arl{0}$), i.e. the average time before a false alarm~\cite{basseville1993}. A well-known manner to control the $\arl{0}$ is to define the thresholds $\{h_t\}_t$ for the test statistics $T_t$ to guarantee that the false alarm probability is constant over time:
% \begin{equation}
%     \mathbb{P}_{\phi_0}(T_t>h_t \mid T_k\leq h_k \forall k<t) = \alpha,
%     \label{eq:alpha}
% \end{equation}
% where $\alpha=1/\arl{0}$~\cite{margavio1995alarm} and $\phi_0$ is the initial data distribution. 
\textbf{Control of the $\arl{0}$.} CDM inherits from QT-EWMA the control of false alarms by maintaining a target $\arl{0}$. In particular, we demonstrate that, since \eqref{eq:alpha} holds for QT-EWMA, CDM yields the same $\arl{0}$ as the QT-EWMA monitoring each class-conditional distribution.
\begin{proposition}
Let $\tilde T$ be the test statistic of CDM defined in \eqref{eq:CDMstat}, and let $\{h_{t}\}$ be the QT-EWMA thresholds 
% used to monitor the class-conditional distributions
yielding the target $\arl{0}$. Then, the change-detection test defined by $\tilde T$ yields the same $\arl{0}$.\label{prop:arl}
\end{proposition}
\begin{proof}
To prove the Proposition, we need to show that \eqref{eq:alpha} holds for $\tilde T$. %The class-specific QT-EWMA statistic $T_t$ is defined as $T^m_{t_m}$, where $m=y_t$, $t_m=\sum_{k=1}^t \mathds{1}(y_k=m)$, and $T^m$ is the QT-EWMA statistic corresponding to the $m$-th class. 
By the definition of $\tilde T$ in \eqref{eq:CDMstat} and the law of total probability we have that
% \begin{equation}
% \begin{aligned}
%     \mathbb{P}_{\phi_0}(&\tilde T_t>\tilde h_t \mid \tilde T_k\leq \tilde h_k\; \forall k<t) = \\
%     =&\sum_{m=1}^M \mathbb{P}_{\phi_0}(T^m_{t_m}>h_{t_m} \mid T^m_k\leq h_k\; \forall k<t_m, y_t=m)\cdot 
%     \\&\qquad\qquad\quad\cdot\mathbb{P}_{\phi_0}(y_t=m \mid T^m_k\leq h_k\; \forall k<t_m).\label{eq:indep}
% \end{aligned}
% \end{equation}
\begin{multline}
    \mathbb{P}_{\phi_0}(\tilde T_t>\tilde h_t \mid \tilde T_k\leq \tilde h_k\; \forall k<t) = \\
    =\sum_{m=1}^M \mathbb{P}_{\phi_0}(T^m_{t_m}>h_{t_m} \mid T^m_k\leq h_k\; \forall k<t_m, y_t=m)\cdot 
    \\\cdot\mathbb{P}_{\phi_0}(y_t=m \mid T^m_k\leq h_k\; \forall k<t_m).\label{eq:indep}
\end{multline}
% \begin{equation}
% \begin{aligned}
%     \mathbb{P}_{\phi_0}(\tilde T_t>\tilde h_t \mid \tilde T_k\leq \tilde h_k\; \forall k<t) = \\
%     =\sum_{m=1}^M \mathbb{P}_{\phi_0}(T^m_{t_m}>h_{t_m} \mid T^m_k\leq h_k\; \forall k<t_m, y_t=m)\cdot 
%     \\\cdot\mathbb{P}_{\phi_0}(y_t=m\midT^m_k\leq h_k\; \forall k<t_m).\label{eq:indep}
% \end{aligned}
% \end{equation}
% \begin{eqnarray}
%     \mathbb{P}_{\phi_0}(\tilde T_t>\tilde h_t \mid \tilde T_k\leq \tilde h_k\; \forall k<t) = \label{eq:fa_prob}\\
%     =\sum_{m=1}^M \mathbb{P}_{\phi_0}(T^m_{t_m}>h_{t_m} \mid T^m_k\leq h_k\; \forall k<t_m, y_t=m)\cdot \label{eq:cond}
%     \\\cdot\mathbb{P}_{\phi_0}(y_t=m\midT^m_k\leq h_k\; \forall k<t_m).\label{eq:indep}
% \end{eqnarray}
Since all the samples $(x_t,y_t)$ in the datastream are assumed to be independent, the label $y_t$ associated with $x_t$ is independent from the values of the statistic $T^m_k$ for $k<t_m$, so we can drop the conditioning in the second factor of the second term of \eqref{eq:indep}. Moreover, the probability under $\phi_0$ in the first factor is conditioned on the event ``$y_t=m$'', so it coincides with the probability under the class-conditional distribution $\phi_0^m$ defined in~\eqref{eq:class_prob}. Hence, \eqref{eq:indep} becomes:
\begin{multline}
    %\mathbb{P}_{\phi_0}(\tilde T_t>\tilde h_t \mid \tilde T_k\leq \tilde h_k\; \forall k<t) = & \\
    \sum_{m=1}^M \mathbb{P}_{\phi_0^m}(T^m_{t_m}>h_{t_m} \mid T^m_k\leq h_k\; \forall k<t_m) % \cdot &\\
    %\cdot
    \mathbb{P}_{\phi_0}(y_t=m) = \\
    =\sum_{m=1}^M\alpha \cdot \mathbb{P}_{\phi_0}(y_t=m) = \alpha,\label{eq:last_eq}
\end{multline}
% \begin{equation}
% \begin{aligned}
%     %\mathbb{P}_{\phi_0}(\tilde T_t>\tilde h_t \mid \tilde T_k\leq \tilde h_k\; \forall k<t) = & \\
%     \sum_{m=1}^M \mathbb{P}_{\phi_0^m}(T^m_{t_m}>h_{t_m} \mid T^m_k\leq h_k\; \forall k<t_m) % \cdot &\\
%     %\cdot
%     \mathbb{P}_{\phi_0}(y_t=m) = & \\
%     =\sum_{m=1}^M\alpha \cdot \mathbb{P}_{\phi_0}(y_t=m) = & \alpha,\label{eq:last_eq}
% \end{aligned}
% \end{equation}
% \begin{eqnarray}
%     %\mathbb{P}_{\phi_0}(\tilde T_t>\tilde h_t \mid \tilde T_k\leq \tilde h_k\; \forall k<t) = & \\
%     \sum_{m=1}^M \mathbb{P}_{\phi_0^m}(T^m_{t_m}>h_{t_m} \mid T^m_k\leq h_k\; \forall k<t_m)\cdot &
%     \\\cdot\mathbb{P}_{\phi_0}(y_t=m) = & \\
%     =\sum_{m=1}^M\alpha \cdot \mathbb{P}_{\phi_0}(y_t=m) = & \alpha,\label{eq:last_eq}
% \end{eqnarray}
where the penultimate equality derives from the fact that \eqref{eq:alpha} holds for the QT-EWMA test statistics $T^m$ monitoring each class-conditional distribution. The last equality in \eqref{eq:last_eq} derives from the assumption that, under $\phi_0$, each sample $x_t$ has a label $y_t=m\in\{1,\ldots,M\}$, so the events $\{y_t=m\}_{m=1}^M$ represent a partition of the probability space, thus their probabilities sum to 1. The fact that \eqref{eq:indep} $=$ \eqref{eq:last_eq} proves that \eqref{eq:alpha} holds for $\tilde T$, showing that CDM yields $\arl{0}=1/\alpha$~\cite{margavio1995alarm}.
\end{proof}

We remark that Proposition \ref{prop:arl} holds for any CDM defined by an online change-detection algorithm that can be configured to yield the desired $\arl{0}$ by setting a constant false alarm probability over time as in \eqref{eq:alpha}. This means that, in principle, we can define CDM using other change-detection tests. However, to the best of our knowledge, QT-EWMA is the only nonparametric and online change-detection test for multivariate datastreams whose thresholds can be set to satisfy \eqref{eq:alpha}, which is not guaranteed by other methods controlling the $\arl{0}$, such as Scan-B~\cite{li2015m} and ECDD~\cite{ross2012exponentially}. %is nonparametric and can be configured to maintain a target $\arl{0}$, but does not guarantee \eqref{eq:alpha}, as experimentally demonstrated in~\cite{frittoli2021change}.

\textbf{Computational Complexity.} Similarly to QT-EWMA~\cite{frittoli2021change}, CDM is extremely efficient in both computational and memory overhead. It places each sample $x_t$ in its bin in the QuantTree histogram $Q^m$ corresponding to its label $m=y_t$, resulting in $\mathcal{O}(K)$ operations~\cite{boracchi2018quanttree}. Then, CDM updates the corresponding statistics $Z^m_{k,t_m}$ \eqref{eq:ewmas} for $k\in\{1,\ldots,K\}$, thus requiring to store in memory $M\cdot K$ values, namely $K$ statistics per class.

\section{Experiments}\label{sec:experiments}
Here we illustrate our experiments, which we designed to demonstrate that CDM outperforms mainstream concept-drift detection methods that monitor either the error rate of a classifier or the overall data distribution. First, we present the real-world and synthetic datasets on which we test our solution (Section~\ref{subsec:datasets}), then we formally define the figures of merit we use (Section~\ref{subsec:figures}) and the reference methods from the literature (Section~\ref{subsec:methods}). Finally, we present and discuss our experiments and their results (Sections~\ref{subsec:exp_real},\ref{subsec:exp_synt}).

%We perform our experiments on synthetic Gaussian datastreams generated by CCM~\cite{carrera2018generating} and on real-world datastreams sampled from the INSECTS dataset~\cite{souza2020challenges}, a well-known concept drift detection benchmark.

\subsection{Considered Datasets}\label{subsec:datasets}

\textbf{Real-world data.} The INSECTS dataset~\cite{souza2020challenges} is a well-known benchmark for classification and concept-drift detection. It contains feature vectors ($d=33$) extracted from sensor measurements describing the wing-beat frequency of six (annotated) species of flying insects. The dataset contains six concepts, each representing measurements acquired at a different temperature, which influences the flying behavior of the insects. This allows us to introduce realistic concept drifts by sampling the datastream from different concepts before and after the change point $\tau$. In our experiments, the stationary condition $\phi_0$ is characterized by the class-conditional distributions $\{\phi_0^m\}_{m=1}^M$ describing the features of $M=4$ different insect species from each of the six concepts. We consider multiple drifts $\phi_0 \to \phi_1$ that consist in a temperature change affecting one or more classes, namely $\phi_0^m \to \phi_1^m \neq \phi_0^m$. In these settings, for each stationary distribution $\phi_0$, the change $\phi_0 \to \phi_1$ is defined among $5$ potential temperature changes affecting one of $2^M-1=15$ different subsets of the $M$ classes, for a total of $75$ distribution changes per initial concept $\phi_0$. In our experiments we consider training sets containing $256$ instances of each class, sampled without replacement from each class-conditional distribution $\phi_0^m$. %In Section~\ref{subsec:exp_real}, we compute the empirical $\arl{0}$ and the detection delay achieved by various models presented in Section~\ref{subsec:methods}.

\begin{figure}
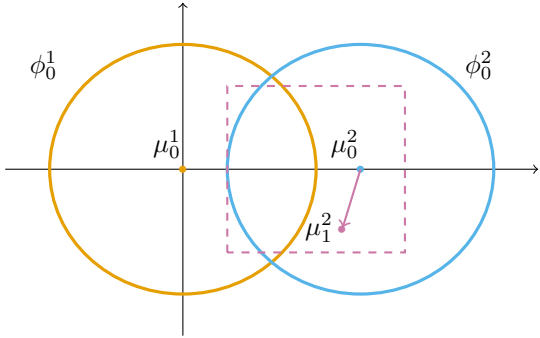

    \centering
    \include{gaussian_setting}
    \vspace{-8mm}
    \caption{
    Illustration of Gaussian class-conditional distributions generating synthetic data. The distributions are represented by the mean and the $3\sigma$ ellipsoid. We consider changes ${\phi_0\to\phi_1}$ defined by translating the mean of $\phi_0^2$ inside the dashed rectangle, as in this example. %Here, an example of change is provided.
    }
    \label{fig:gaussian_setting}
\end{figure}

\textbf{Synthetic data.} To interpret the results obtained on real-world data, we synthetically generate various distribution changes $\phi_0 \to \phi_1$ and assess their impact on the classification error. In particular, we define the stationary distribution $\phi_0$ as a mixture of $M=2$ Gaussians (one per class) $\phi_0^1=\mathcal{N}(\mu_0^1,I)$ and $\phi_0^2=\mathcal{N}(\mu_0^2,I)$ in $\mathbb{R}^2$, where $I$ denotes the identity matrix, $\mu_0^1 = [0,0]^T$, and $\mu_0^2=[\delta, 0]^T$ for some $\delta > 0$. Post-change distribution $\phi_1$ is defined by shifting $\phi_0^2 \to \phi_1^2=\mathcal{N}(\mu_1^2,I)$, while keeping $\phi_0^1$ fixed. Changes are thus regulated by $\mu_1^2$, which we move over a grid around $\mu_0^1$ (see Figure~\ref{fig:gaussian_setting}). Also in this case, we consider training sets containing $256$ samples drawn from each $\phi_0^m$. 

This setup was designed to assess when CDM is a better option than ECDD. The classification error varies when $\mu_1^2$ moves along the horizontal direction, which is the line connecting $\mu_0^1$ and $\mu_0^2$: these changes can be promptly detected by ECDD when they increase the error rate. In contrast, changes translating $\mu_1^2$ vertically (thus orthogonal to the line joining $\mu_0^1$ and $\mu_0^2$), do not change the error rate but only the input distribution. These changes cannot be detected by ECDD, but are perceivable by CDM, whose performance only depends by the change magnitude. Here we measure the change magnitude by the symmetric Kullback-Leibler distance $sKL(\phi_0^2, \phi_1^2)$~\cite{kullback1951information}, which in this case is equal to $\frac{1}{2}\Vert\mu_1^2-\mu_0^2\Vert_2$.

\subsection{Figures of Merit}\label{subsec:figures}
We consider two common figures of merit in the change-detection literature. First, we assess the control of false alarms by computing the \emph{empirical} $\arl{0}$, i.e., the average detection time in datastreams distributed as $\phi_0$. Thanks to Proposition~\ref{prop:arl}, we expect the empirical $\arl{0}$ of CDM to approach the target $\arl{0}$ set before monitoring. Then, we measure the detection power by the \emph{average detection delay} (or $\arl{1}$), namely the average difference between the detection time $t^*$ and the actual change point $\tau$. The detection delay is computed considering only datastreams where no false alarms were reported before the change, thus $t^*>\tau$.

% After configuring all the methods to yield similar values of empirical $\arl{0}$, we adopt the procedure in in~\cite{demvsar2006statistical} to compare the detection power. In each experiment on the INSECTS dataset we rank the considered methods according to their average detection delay (rank $=1$ for the method with the lowest detection delay, etc.), and compute their average rank over multiple changes and iterations. Finally, we perform the Nemenyi~\cite{nemenyi1963distribution} and Dunn~\cite{dunn1961multiple} post-hoc tests to assess whether the performance difference between the best-ranking method and the others is statistically significant.

% We also analyze the computational complexity and memory requirements of all the tested methods and report the comparison in Table~\ref{tab:complexity}

% \lu{Questa parte forse si può spostare dopo, non è proprio una cifra di merito.} \di{Essendo una cosa che misuriamo, la terre in questa sottosezione, ma concordo col fatto che non sia una figures of merit che descrive le performance del metodo.} Finally, in every experiment, we measure the classification error rate before and after the change, respectively denoted with $p_0$ and $p_1$. % These values will allow us to interpret some of the results from our experiments.

\subsection{Considered Methods}\label{subsec:methods}
% To prove that the proposed class-specific monitoring is superior to the classic class-independent monitoring, we consider two nonparametric models, namely Scan-B~\cite{li2015m} and QT-EWMA~\cite{frittoli2021change}. In addition to these, we employ ECDD~\cite{ross2012exponentially}, which monitors the evolution of the classification error via an exponentially weighted moving average.

% In the following, we describe how we set the considered models to ensure a fair comparison. Since a larger $\arl{0}$ corresponds to a slower detection, we only considered models that have been designed to yield a target $\arl{0}$ (equivalently, False Positive Rate). However, the empirical $\arl{0}$ achieved by some of these models does not match their target $\arl{0}$. Hence, we decided to set the target $\arl{0}$ of the models to match their empirical $\arl{0}$. In Section~\ref{subsec:arl0}, we present the results showing the empirical $\arl{0}$ achieved by the considered models, and supporting our setup choices.

% Here we present the concept-drift detection methods against which we compare our proposed solution. 
To ensure a fair comparison, we only consider methods that \emph{i)} are nonparametric and \emph{ii)} control the false alarms by setting a target $\arl{0}$ before monitoring. In particular, we consider ECDD~\cite{ross2012exponentially}, which monitors the error rate of a classifier, Scan-B~\cite{li2015m} and QT-EWMA~\cite{frittoli2021change}, which are nonparametric and online change-detection tests monitoring the data distribution.

%\begin{itemize}
    %\item 
\textbf{ECDD} \cite{ross2012exponentially} employs an EWMA control chart~\cite{roberts1959control} to monitor the sequence $\{e_t\}$ \eqref{eq:errors} defined by the errors of a classifier $\mathcal{K}$. In particular, ECDD computes a statistic
\begin{equation}
    U_t = (1-r)U_{t-1} + re_t,\quad U_0 = \hat p_{0,0},\label{eq:ecdd}
\end{equation}
where $\hat p_{0,t}$ indicates the average error rate of $\mathcal{K}$ up to time $t$, and $r$ is the EWMA parameter, which we set to $r=0.2$ as in \cite{ross2012exponentially}. A concept drift is detected when $U_t>\hat p_{0,t}+L\sigma_t$, where $\sigma_t$ is the estimated standard deviation of $U_t$:
\begin{equation}
    \sigma_t = \sqrt{\hat p_{0,t}(1-\hat p_{0,t})\dfrac{r}{2-r}(1-(1-r)^{2t})}.
\end{equation}
Since $U_t$ is an incremental estimate of the error rate of $\mathcal{K}$, which gives exponentially larger weights to the latest elements of $e_t$ compared to older elements, and since a one-sided decision rule is applied, ECDD can only detect drifts that increase the classification error. The control limit $L$ can be tuned to yield a target $\arl{0}$, and \cite{ross2012exponentially} provides polynomial approximations to compute $L$ for different values of the target $\arl{0}$ as a function of $\hat p_{0,t}$.

In our experiments on INSECTS, we train $\mathcal{K}$ as a $k$-Nearest Neighbors ($k$-NN) classifier ($k=9$), and we never update it during monitoring. On the synthetic dataset we employ a Linear Discriminant Analysis (LDA) classifier, which is faster and yields excellent performance over Gaussian classes. 

In terms of computational complexity, computing and updating $U_t$ \eqref{eq:ecdd} and $\hat p_{0,t}$ are extremely cheap operations, which require storing in memory only 2 scalar values, namely $U_{t-1}$ and $\hat p_{0,t-1}$. The computational complexity of ECDD therefore depends on that of the classifier, which we indicate by $\mathcal{O}(\mathcal{K})$, which has to be applied on each $x_t$. 

\input{tables/table_complexity}

\textbf{Scan-B} \cite{li2015m} is a nonparametric change-detection algorithm that monitors the input distribution by computing at each time $t$, the average Maximum Mean Discrepancy (MMD)~\cite{gretton2012kernel} between a sliding window of a fixed size $B$ and $n$ reference windows of the same size sampled from the training set $TR$. This requires updating $n$ Gram matrices for each sample $x_t$ by computing $B$ times the MMD statistic, resulting in $\mathcal{O}(nBd)$ operations~\cite{keriven2020newma}. Therefore, Scan-B stores in memory $n$ reference windows of $B$ $d$-dimensional inputs, on top of the current window, resulting in $(n+1)Bd$ memory footprint. Thresholds are set by analyzing the asymptotic behavior of $\arl{0}$ when the threshold tends to infinity, while in CDM and QT-EWMA the thresholds are defined by \eqref{eq:alpha}, providing more accurate control of the $\arl{0}$~\cite{frittoli2021change}. %rather than by \eqref{eq:alpha} \gi{non chiaro questo rather than... e poi mettiamo le mani avanti per dire che?}\di{Credo si voglia dire che non lo mantiene con indipendenza dal tempo? Ma anche a me non risulta chiarissimo.}. 
As in \cite{li2015m}, we set the window size $B=50$ and $n=5$.
    %\item 
    
\textbf{QT-EWMA} \cite{frittoli2021change}, is a nonparametric change-detection algorithm which we have described in Section~\ref{subsec:detectors}. To enable a fair comparison, since CDM leverages $M$ instances of QT-EWMA  each one based on a QuantTree histogram with $K=16$ bins, we set the number of bins of QT-EWMA to $MK$ and, according to ~\cite{frittoli2021change}, we set $\lambda=0.03$ in \eqref{eq:ewmas}. As shown in~\cite{frittoli2021change}, QT-EWMA is very efficient since it performs $\mathcal{O}(MK)$ operations to place each sample $x_t$ in the corresponding bin of the QuantTree histogram~\cite{boracchi2018quanttree}, and requires storing only the $MK$ scalar values of the statistic $Z_{k,t-1}$ \eqref{eq:ewmas} to be updated at time $t$.

In Table \ref{tab:complexity} we compare the computational complexity and memory requirements of CDM (discussed in Section \ref{subsec:arl}) to those of the other considered methods. This analysis shows that CDM and QT-EWMA are extremely efficient from both the computational and memory points of view. In contrast, Scan-B performs more operations and stores more data, and these requirements increase with the data dimension $d$, contrarily to CDM and QT-EWMA. ECDD has negligible memory requirements, but its computational complexity depends on the classifier $\mathcal{K}$, which is applied to each sample $x_t$ to form $\{e_t\}$.
% \gi{La commentiamo con qualche confronto diretto da qualche parte nel testo?}

% \gi{dei training set non parliamo mai?}
% \end{itemize}

% \subsubsection{ECDD}
% \begin{itemize}
%     \item target $\arl{0}=400$
%     \item $\lambda = 0.2$
% \end{itemize}

% \subsubsection{Scan-B}
% \begin{itemize}
%     \item target $\arl{0}=300$
%     \item $B_0 = 0.2$ \lu{non era 50?}
%     \item $n=5$
% \end{itemize}

% \subsubsection{QT-EWMA}
% Class independent:
% \begin{itemize}
%     \item target $\arl{0}=375$
%     \item $K = 16*M$
%     \item $\lambda = 0.03$
% \end{itemize}

% CDM:
% \begin{itemize}
%     \item target $\arl{0}=375$
%     \item $K = 16$
%     \item $\lambda = 0.03$
% \end{itemize}

% \subsubsection{Hybrid Monitoring}
% Some details...
% The class-specific version...

\input{tables/table_insects_arl0}

\subsection{Concept Drift Detection on INSECTS Data}\label{subsec:exp_real}
In this Section, we discuss the empirical $\arl{0}$ and the detection delay achieved on the INSECTS dataset by the considered models in the settings described in Section~\ref{subsec:datasets}.
% In these experiments, we consider $6$ potential stationary distributions $\phi_0$, one for each concept (i.e., temperature) available in the dataset. Then, the distribution change $\phi_0 \to \phi_1$ is defined as a one of the $5$ potential temperature changes from $\phi_0$, for a total of $30$ different concept changes. Moreover, for each concept change, we consider all the $2^4-1=15$ subsets of insects species that can be affected by the change, for a total of $450$ potential changes. In the following, we report and discuss the results achieved by the considered models, set up as in Section~\ref{subsec:methods}.

% ARL0: for each phi_0 we repeat the experiment to compute the ARL0
\textbf{$\boldarl{0}$.} We compute the empirical $\arl{0}$ of the considered methods on the six concepts of the INSECTS dataset~\cite{souza2020challenges}, which we denote by A, B, C, D, E, F. We consider each concept as a stationary distribution $\phi_0$, and we sample without replacement %\gi{con reinserimento?}\di{nessun reinserimento in un datastream, ma reinseriamo tutto prima del nuovo datastream. Quindi "no reinseriimento" direi.} 
$5000$ training sets and $5000$ datastreams of length $8000$ from each $\phi_0$. %\gi{la lunghezza del TR l'abbiamo detta da qualche parte? del test set?}. 
Then, we configure the considered methods on the training sets, and compute the empirical $\arl{0}$ as the average detection time over these stationary datastreams. 

We report the results of this experiment in Table~\ref{tab:insects_arl0}, which shows that ECDD fails at accurately controlling the target $\arl{0}=400$. In contrast, the empirical $\arl{0}$ of CDM and QT-EWMA approaches their target, which we set to $\arl{0}=375$ to match the empirical $\arl{0}$ of ECDD. Similarly to ECDD, Scan-B does not accurately control the $\arl{0}$, and this is consistent with the experiments in~\cite{frittoli2021change}. For this reason, we set the target $\arl{0}=300$ in Scan-B to yield approximately the same empirical $\arl{0}$ as the other methods. Table~\ref{tab:insects_arl0} indicates that in these settings it is possible to fairly compare the detection delays of the considered methods, since these all yield approximately the same empirical $\arl{0}$. %have been correctly set up to yield the same empirical $\arl{0}$, even though their target is different ($400$ and $300$ respectively). This experiment ensures that the comparison between the considered models with their respective setup is fair and statistically meaningful.

\input{tables/table_insects_delay_species}

% When computing the average detection delay for a model, we only consider those streams where no false alarms were reported before $\tau$, thus $t^* > \tau$.
% over $1000$ datastreams per initial concept, depending on the drifted classes. % achieved by the considered models averaged over $1000$ streams for each setting, grouped and average according to different criteria. 
\textbf{Detection delay.} For each of the $450$ changes $\phi_0 \to \phi_1$ ($75$ for each of the $6$ initial concepts) described in Section~\ref{subsec:datasets}, we sample without replacement $1000$ training sets and $1000$ datastreams to be monitored. Each datastream is the concatenation of $\tau=160$ points drawn from $\phi_0$ and $7000$ points drawn from $\phi_1$. Table~\ref{tab:insects_delays_species} reports the average detection delays of the considered methods depending on the drifted classes. As suggested in~\cite{demvsar2006statistical}, we rank the considered methods according to their average detection delay obtained on each of the $450$ changes (rank $=1$ for the method with the lowest detection delay, etc.), and report their average rank. We also report the p-values of the Nemenyi~\cite{nemenyi1963distribution} and Dunn~\cite{dunn1961multiple} post-hoc tests, to assess whether the differences between the best-ranking method and the others are statistically significant. %\di{Ripetizione di Nem/Dunn da V-B?}

% We compute the detection delay $t^* - \tau$ achieved by the considered models for every change $\phi_0 \to \phi_1$, where we also consider different subsets of classes affected by the change. For each change $\phi_0 \to \phi_1$ (concept change, subset of classes) we sample from the INSECTS dataset $1000$ 

% 1. Noi siamo meglio degli altri quasi sempre (ed è statisticamente valida come differenza)
We observe that CDM turns out to be the best method in 13 out of the 15 considered changes, and the best in terms of average rank. The Nemenyi and Dunn tests show that the gap of CDM over ECDD and QT-EWMA is statistically significant (p-value $<0.05$). The gap between CDM and Scan-B is less remarkable, but still significant according to the Dunn test. 

% 2. Tutti i metodi migliorano quando più classi sono colpite dal change (spiegazione!) In particolare, guardiamo CDM vs QT-EWMA!
As expected, all the methods tend to yield lower detection delays when the change affects more classes. %In fact, when all the classes are affected by the change, every incoming sample weighs on the models' statistics towards the detection. 
In particular, the difference between the detection delays of CDM and QT-EWMA is larger when the change affects only one class rather than when it affects all of them, showing that monitoring the class-conditional distributions can indeed improve the detection performance in these cases.
% 3. Inoltre, se guardiamo CDM vs Scan-B
This effect is even more apparent in the comparison between CDM and Scan-B. %, where the latter performs slightly better when the change affects all the classes.

\begin{figure*}[t!]
    \centering
    \includegraphics[width=\textwidth]{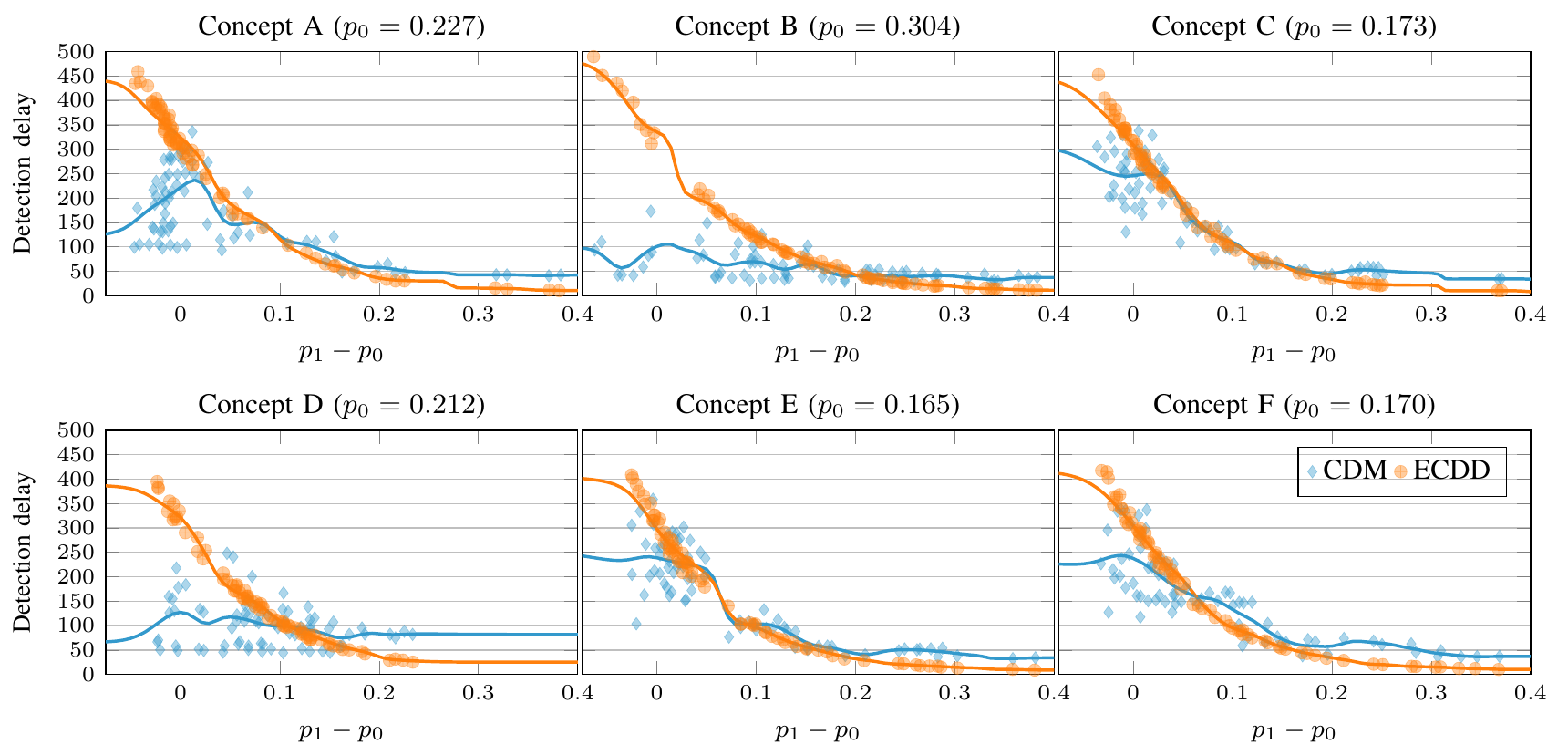}
    \caption{
    Detection delay achieved by ECDD and CDM on the INSECTS dataset~\cite{souza2020challenges} for each of the $6$ stationary concepts, plotted against the difference $p_1-p_0$, where $p_0,p_1$ are the error rates of $\mathcal{K}$ before and after the drift. Each dot is the average of $1000$ realizations of the same change $\phi_0 \to \phi_1$ i.e., thus with the same affected classes.
    }
    \label{fig:delay_insects}
\end{figure*}

% 4. Infine, notiamo come ECDD non riesca a competere (in media) ma riesce a batterci quando deltap0 è alto (grafici)
% Conclusioni: non tutti i change reali impattano la classificazione, andrebbero analizzati
%Finally, ECDD proves to be unable to consistently and promptly detect changes on the INSECTS dataset, as this method specifically monitors the unilateral divergence of the classification error. However, the considered INSECTS dataset shows that there exist changes that might not affect significantly the performance of a classifier, hence rendering sequential methods like ECDD powerless. 
Most remarkably, CDM substantially outperforms ECDD in terms of average detection delay in all the considered settings. This is due to the fact that ECDD can only detect concept drifts that increase the classification error, while the considered drifts in the INSECTS dataset might have little impact on the error rate of a classifier. To further analyze the relation between detection delay and classification error, we plot in Figure~\ref{fig:delay_insects} the average detection delays of CDM and ECDD against the difference between the classification error after ($p_1$) and before the change ($p_0$). Each plot reports the results obtained on the $75$ drifts we consider for each initial concept A, B, C, D, E, F. We highlight the relation between $p_1-p_0$ and the detection delay by plotting the moving average (weighted by a Gaussian kernel) of the detection delay as a function of $p_1-p_0$. 
%Each dot represents the average detection delay achieved on the $1000$ datastreams generated in the same $\phi_0 \to \phi_1$ setting. 
These results qualitatively show that the performance of ECDD only depends on $p_1 - p_0$, which is often small and sometimes even negative. In contrast, CDM can detect any change in the class-conditional distributions, thus yielding a lower detection delay in most cases. %\gi{dire come hai ottenuto le curve}\di{Each dot...}

\subsection{Concept Drift Detection on Gaussian Data}\label{subsec:exp_synt}
% The results on the INSECTS dataset discussed in the previous section hint to the fact that classification-error based models cannot detect a series of changes that only affect the data distribution. In this section, we consider a simplified synthetic scenario where we can control the impact of the change over the classification error, gaining insight on the relation between this and the performance of CDM and ECDD.

\begin{figure*}[t!]
    \centering
    \includegraphics[width=.9\textwidth]{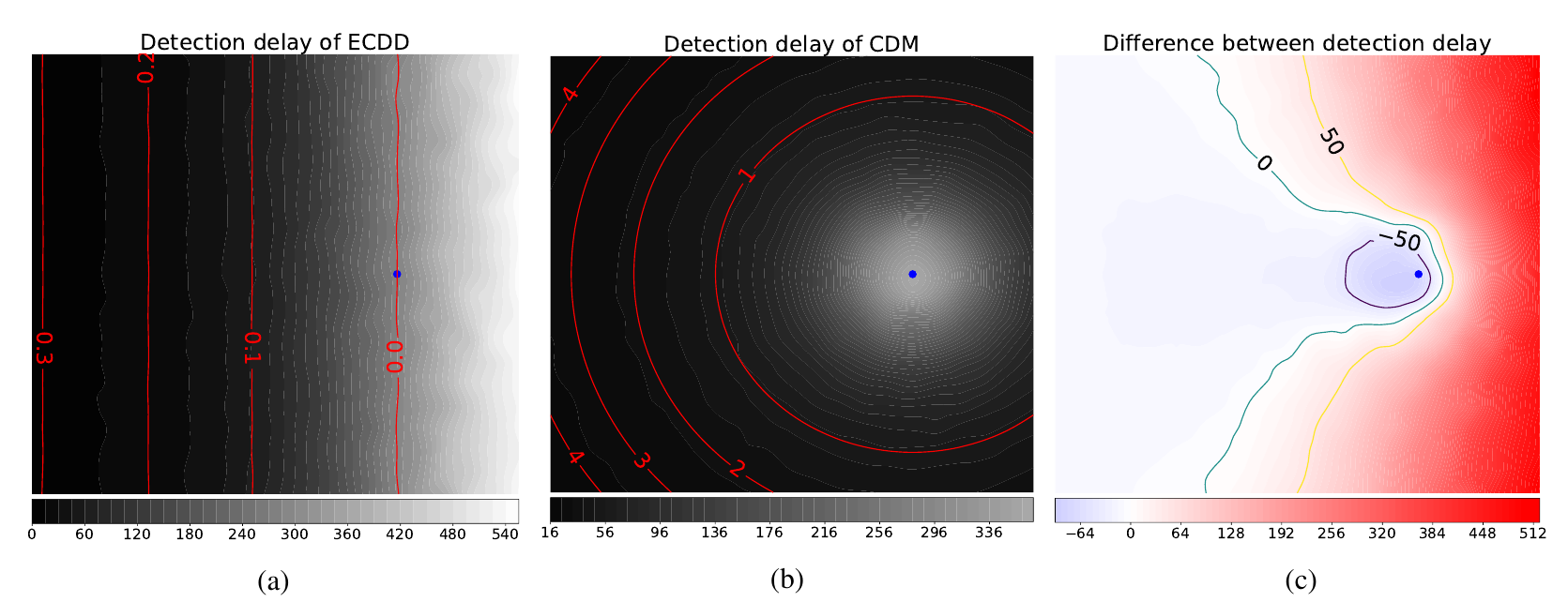}
    \caption{
    Results of the experiment on synthetic data. The blue dot represents the pre-change mean $\mu_0^2$. (a,b) report, at each coordinate $\overline{\mu}$, the average detection delay achieved by ECDD and CDM, respectively, when $\mu_0^2 \to \overline{\mu}$. In (a) the contour lines indicate the difference in classification error $p_1 - p_0$ before and after the change, and in (b) the change magnitude $\skl(\phi_0^2, \phi_1^2)$. (c) report the difference between the detection delay achieved over synthetic data by ECDD and CDM, with contour lines.
    }
    \label{fig:heatmaps}
\end{figure*}

% Intro. I change non impattano allo stesso modo sul classification error (lo abbiamo visto nella sezione precedente). Qui vogliamo analizzare la relazione tra detection delay, classification error e KL distance per mostrare che ECDD-like models non sono generali.
% \gi{dire come hai ottenuto le immagini? Forse diventa piu\' facile partire dai claims}
% As we observed in Section~\ref{subsec:exp_real}, concept drifts  might not always heavily impact the classification performance \gi{dove lo si vede? aiutiamo il lettore}.
Concept drifts might not always heavily impact the classification performance, as we have shown in Figure~\ref{fig:delay_insects} on the INSECTS dataset. Here we further analyze the fundamental difference between monitoring the classification error (ECDD) and the input distribution (CDM), by considering the synthetic scenario described in Section~\ref{subsec:datasets}, where we can control both $p_1 - p_0$ and the change magnitude $\skl(\phi_0^2, \phi_1^2)$. We configure ECDD and CDM to maintain the same $\arl{0}$ as in Section~\ref{subsec:exp_real}.
The results of this experiment are illustrated in Figure~\ref{fig:heatmaps}. Figures~\ref{fig:heatmaps}(a,b) report the detection delays respectively achieved by ECDD and CDM as a heatmap. The color coded value at a coordinate $\overline{\mu} \in \mathbb{R}^2$ represents the detection delay achieved by the model when $\mu_1^2 = \overline{\mu}$, averaged over $5000$ experiments. Moreover, in the same figures we report, respectively, $p_1 - p_0$ and $\skl(\phi_0^2, \phi_1^2)$ as contour plots. 

As expected, ECDD cannot detect virtual drifts, as can be seen by the large detection delays on the right side of Figure~\ref{fig:heatmaps}(a), but it achieves excellent detection performance when the translation reduces the distance between the two class-conditional distributions, increasing the classification error ($p_1 - p_0 > 0$). In contrast, the detection delay of our CDM only depends on the distance $\skl(\phi_0^2, \phi_1^2)$, as it can be appreciated in Figure~\ref{fig:heatmaps}(b), where the level curves of the detection delays are circular and follow $\frac{1}{2}\Vert\mu_1^2-\mu_0^2\Vert_2$.
Figure~\ref{fig:heatmaps}(c) reports the difference between the detection delays of ECDD and CDM. %, together with the contour lines for the values $-50$, $0$ and $50$. It is evident that ECDD is limited to the detection of changes in a specific direction, rather than generic distribution changes. 
ECDD outperforms CDM when $\mu_1^2$ falls inside a relatively small triangular portion of $\mathbb{R}^2$, corresponding to drifts that significantly increase the error rate while keeping the distance between $\phi_0^2$ and $\phi_1^2$ low. However, the difference is substantial only in a small region close to $\mu_0^2$, where the change is nearly negligible and the performance of both methods is rather poor. %However, the performance difference between ECDD and CDM is not substantial  Finally, 
CDM yields lower detection delays than ECDD in all the other cases, and the performance difference is quite large, especially when the drift reduces the classification error.

\section{Conclusions and Future Work}
We have introduced CDM, a novel concept-drift detection method that monitors the class-conditional distributions using QT-EWMA~\cite{frittoli2021change}. Our experiments on real-world datastreams and synthetic data show that our solution can effectively detect virtual drifts that are ignored by methods that monitor the error rate of a classifier. In many circumstances, CDM  yields lower detection delays than methods that monitor the overall data distribution, especially when the drift affects only a subset of classes. Moreover, our CDM is built upon solid theoretical guarantees on false alarms, enabling to set the $\arl{0}$ before monitoring. Another important advantage of CDM compared to other concept-drift detection methods is that our solution returns information on which class triggered the detection, and this can be crucial for adaptation and diagnostics in general. 

Future work will extend CDM by applying other detectors controlling the $\arl{0}$ in parametric settings or in combination with QT-EWMA, to improve the detection performance when parametric assumptions can be made on some class-conditional distributions. We will also investigate the application of CDM when class labels are not available during monitoring, using the predictions of a classifier instead. %of the actual class labels. %\gi{classifier to form the stream?}

\bibliographystyle{IEEEtran}
\bibliography{biblio}

\end{document}

%% file: figures/setup_tikz.tex
\definecolor{cdm_c}{rgb}{.2,.6,.8}
\definecolor{ecdd_c}{rgb}{1,.5,.05}
% \definecolor{cdm_c}{HTML}{2FABFB}
% \definecolor{ecdd_c}{HTML}{FB3F2F}
\definecolor{qtewma_c}{HTML}{117733}
\definecolor{scanb_c}{HTML}{44AA99}

\definecolor{colorblind1}{HTML}{000000}
\definecolor{colorblind2}{HTML}{E69F00}
\definecolor{colorblind3}{HTML}{56B4E9}
\definecolor{colorblind4}{HTML}{009E73}
\definecolor{colorblind5}{HTML}{F0E442}
\definecolor{colorblind6}{HTML}{0072B2}
\definecolor{colorblind7}{HTML}{D55E00}
\definecolor{colorblind8}{HTML}{CC79A7}

% \definecolor{our_blue}{rgb}{.2,.6,.8}
% \definecolor{our_orange}{rgb}{1,.5,.05}
% https://htmlcolorcodes.com/

% \definecolor{spll_c}{HTML}{999933}
% \definecolor{pcaspll_c}{HTML}{DDCC77}
% \definecolor{density_tree_c}{HTML}{CC6677}

\newcommand{\cdmMarker}{diamond*}
\newcommand{\qtewmaMarker}{pentagon*}
\newcommand{\ecddMarker}{oplus*}
\newcommand{\scanbMarker}{oplus}
\newcommand{\LEGEND}[1]{\addlegendentry{#1}}

%% file: gaussian_setting.tex
\begin{tikzpicture}
    \begin{axis}[
        width=0.8\columnwidth,
        height=0.5\columnwidth,
        scale only axis,
        xmin=-2, xmax=4,
        ymin=-2, ymax=2,
        axis line style={draw=none},
        tick style={draw=none},
        xticklabels={,,},
        yticklabels={,,}
        ]
    %,
    \draw[color=colorblind1, thin, ->](-2,0) -- (4,0);
    \draw[color=colorblind1, thin, ->](0,-2) -- (0,2);
    \filldraw[color=colorblind2, fill=none, very thick](0,0) circle (1.5);
    \node [colorblind2] at (0,-.035) {\textbullet};
    \node[text width=1cm] at (0.09,0.3) {$\mu_0^1$};
    \node[text width=1cm] at (-1.3,1.25) {$\phi_0^1$};
    \filldraw[color=colorblind3, fill=none, very thick](2,0) circle (1.5);
    \node [colorblind3] at (2,-.035) {\textbullet};
    \node[text width=1cm] at (2.09,0.3) {$\mu_0^2$};
    \node[text width=1cm] at (3.6,1.25) {$\phi_0^2$};
    \draw[color=colorblind8, thick, ->](2,0) -- (1.8,-0.7);
    \node[text width=1cm] at (1.8,-0.7) {$\mu_1^2$};
    \node [colorblind8] at (1.79,-0.75) {\textbullet};
    \draw[colorblind8, thick, dashed] (0.5,-1.0) -- (0.5,1.0) -- (2.5,1.0) -- (2.5,-1.0) -- cycle;
    \end{axis}
\end{tikzpicture}

%% file: tables/table_complexity.tex
\begin{table}[t!]
\centering
\caption{Computational complexity for processing a new sample $(x_t,y_t)$ and memory requirement of CDM and the other considered methods. 
%\di{In our experiments we set the number of bins $K=16$ for CDM and $MK$ for QT-EWMA~\cite{frittoli2021change}, and $B=50, n=5$ for Scan-B~\cite{li2015m}.} 
The computational complexity of ECDD~\cite{ross2012exponentially} is that of the classifier $\mathcal{K}$, indicated by $\mathcal{O}(\mathcal{K})$.
}\label{tab:complexity}
\resizebox{\columnwidth}{!}{  
\begin{tabular}{c|cccc} 
\hline
Method & ECDD~\cite{ross2012exponentially} & Scan-B~\cite{li2015m} & QT-EWMA~\cite{frittoli2021change} & CDM (ours)\\
\hline
Complexity & $\mathcal{O}(\mathcal{K})$ & $\mathcal{O}(nBd)$ & $\mathcal{O}(MK)$ & $\mathcal{O}(K)$ \\
Memory & $2$ & $(n+1)Bd$ & $MK$ & $MK$ \\
\hline
\end{tabular}
}
\end{table}

%% file: tables/table_insects_arl0.tex
\begin{table}[t!]
\centering
\caption{Empirical $\arl{0}$ of the considered methods on the 6 concepts of the INSECTS dataset~\cite{souza2020challenges}.
% Despite having different targets, the considered methods achieve a similar empirical $\arl{0}$.
% We configured all methods to maintain approximately the same $\arl{0}=375$. Since ECDD and Scan-B do not accurately maintain the desired $\arl{0}$, for these methods we set the target ${\arl{0}=400,300}$, respectively.
%For each concept we report the error rate $p_0$ of the classifier.
}\label{tab:insects_arl0}
\resizebox{\columnwidth}{!}{  
\begin{tabular}{c|cccc} 
\hline
 & \multicolumn{4}{c}{Method (target $\arl{0}$)}\\
 \cline{2-5}
Concept & ECDD~\cite{ross2012exponentially} & Scan-B~\cite{li2015m} & QT-EWMA~\cite{frittoli2021change} & CDM (ours)\\
% Concept & ECDD [4] & Scan-B [5] & QT-EWMA [2] & CDM (ours)\\
 & ($400$) & ($300$) & ($375$) & ($375$) \\
\hline
A & $376.51$ & $382.08$ & $379.10$ & $375.44$ \\
B & $371.07$ & $384.56$ & $361.78$ & $374.47$ \\
C & $373.16$ & $381.65$ & $371.66$ & $365.32$ \\
D & $374.14$ & $387.17$ & $367.18$ & $369.94$ \\
E & $371.82$ & $376.28$ & $375.10$ & $374.64$ \\
F & $377.67$ & $374.22$ & $375.58$ & $371.87$ \\
% 1 ($0.227$) & $376.51$ & $382.08$ & $379.10$ & $375.44$ \\
% 2 ($0.304$) & $371.07$ & $384.56$ & $361.78$ & $374.47$ \\
% 3 ($0.173$) & $373.16$ & $381.65$ & $371.66$ & $365.32$ \\
% 4 ($0.212$) & $374.14$ & $387.17$ & $367.18$ & $369.94$ \\
% 5 ($0.165$) & $371.82$ & $376.28$ & $375.10$ & $374.64$ \\
% 6 ($0.170$) & $377.67$ & $374.22$ & $375.58$ & $371.87$ \\
\hline
\end{tabular}
}
\end{table}

%% file: tables/table_insects_delay_species.tex
\begin{table}[t!]
\centering
\caption{Average detection delays on the 15 subsets of classes affected by change of the INSECTS dataset~\cite{souza2020challenges}.
}\label{tab:insects_delays_species}
\resizebox{\columnwidth}{!}{  
\begin{tabular}{c|cccc} 
\hline
Drifted classes & ECDD~\cite{ross2012exponentially} & Scan-B~\cite{li2015m} & QT-EWMA~\cite{frittoli2021change} & CDM (ours)\\
\hline
1          & $207.98$ & $212.53$ & $267.73$ & $\mathbf{195.45}$ \\  
2          & $245.85$ & $162.58$ & $195.44$ & $\mathbf{124.92}$ \\  
3          & $264.27$ & $224.99$ & $278.57$ & $\mathbf{204.00}$ \\  
4          & $224.91$ & $235.87$ & $265.96$ & $\mathbf{196.74}$ \\
\hline
1,2        & $198.17$ & $131.71$ & $174.80$ & $\mathbf{114.44}$ \\  
1,3        & $172.62$ & $169.87$ & $223.50$ & $\mathbf{160.98}$ \\  
1,4        & $165.77$ & $163.63$ & $221.66$ & $\mathbf{145.82}$ \\  
2,3        & $163.66$ & $126.56$ & $167.55$ & $\mathbf{112.18}$ \\  
2,4        & $176.53$ & $119.41$ & $154.95$ & $\mathbf{106.49}$ \\  
3,4        & $210.04$ & $169.88$ & $218.90$ & $\mathbf{153.51}$ \\  
\hline
1,2,3      & $139.29$ & $115.01$ & $152.91$ & $\mathbf{103.60}$ \\  
1,2,4      & $148.03$ & $103.24$ & $141.09$ & $\mathbf{98.89}$  \\  
1,3,4      & $144.81$ & $134.83$ & $183.41$ & $\mathbf{131.38}$ \\  
2,3,4      & $132.36$ & $\mathbf{96.92}$  & $136.90$ & $98.57$  \\  
\hline
1,2,3,4    & $122.38$ & $\mathbf{88.86}$  & $128.04$ & $91.44$  \\  
% \hhline{=====}
\hline
\hline
Avg. rank       & $2.416$  & $2.356$  & $3.524$  & $\mathbf{1.704}$           \\ 
Nemenyi-p      & $6.95\cdot10^{-6}$ & $1.44\cdot10^{-1}$ & $5.18\cdot10^{-17}$         & -- \\  
Dunn-p          & $2.43\cdot10^{-7}$ & $2.01\cdot10^{-2}$ & $6.40\cdot10^{-19}$       & -- \\ 
\hline
\end{tabular}
}
\end{table}

%% file: main.bbl
\begin{thebibliography}{10}
\providecommand{\url}[1]{#1}
\csname url@samestyle\endcsname
\providecommand{\newblock}{\relax}
\providecommand{\bibinfo}[2]{#2}
\providecommand{\BIBentrySTDinterwordspacing}{\spaceskip=0pt\relax}
\providecommand{\BIBentryALTinterwordstretchfactor}{4}
\providecommand{\BIBentryALTinterwordspacing}{\spaceskip=\fontdimen2\font plus
\BIBentryALTinterwordstretchfactor\fontdimen3\font minus
  \fontdimen4\font\relax}
\providecommand{\BIBforeignlanguage}[2]{{%
\expandafter\ifx\csname l@#1\endcsname\relax
\typeout{** WARNING: IEEEtran.bst: No hyphenation pattern has been}%
\typeout{** loaded for the language `#1'. Using the pattern for}%
\typeout{** the default language instead.}%
\else
\language=\csname l@#1\endcsname
\fi
#2}}
\providecommand{\BIBdecl}{\relax}
\BIBdecl

\bibitem{bahri2021data}
M.~Bahri, A.~Bifet, J.~Gama, H.~M. Gomes, and S.~Maniu, ``Data stream analysis:
  Foundations, major tasks and tools,'' \emph{WIREs: Data Mining and Knowledge
  Discovery}, vol.~11, no.~3, p. e1405, 2021.

\bibitem{lu2018learning}
J.~Lu, A.~Liu, F.~Dong, F.~Gu, J.~Gama, and G.~Zhang, ``Learning under concept
  drift: A review,'' \emph{IEEE Transactions on Knowledge and Data
  Engineering}, vol.~31, no.~12, pp. 2346--2363, 2018.

\bibitem{geng2020recent}
C.~Geng, S.-J. Huang, and S.~Chen, ``Recent advances in open set recognition: A
  survey,'' \emph{IEEE Transactions on Pattern Analysis and Machine
  Intelligence}, vol.~43, no.~10, pp. 3614--3631, 2021.

\bibitem{basseville1993}
M.~Basseville, I.~V. Nikiforov \emph{et~al.}, \emph{Detection of abrupt
  changes: theory and application}.\hskip 1em plus 0.5em minus 0.4em\relax
  Prentice Hall Englewood Cliffs, 1993, vol. 104.

\bibitem{frittoli2021change}
L.~Frittoli, D.~Carrera, and G.~Boracchi, ``Change detection in multivariate
  datastreams controlling false alarms,'' in \emph{Joint European Conference on
  Machine Learning and Knowledge Discovery in Databases}.\hskip 1em plus 0.5em
  minus 0.4em\relax Springer, 2021, pp. 421--436.

\bibitem{boracchi2018quanttree}
G.~Boracchi, D.~Carrera, C.~Cervellera, and D.~Macci\`o, ``Quant{T}ree:
  histograms for change detection in multivariate data streams,'' in
  \emph{International Conference on Machine Learning}.\hskip 1em plus 0.5em
  minus 0.4em\relax PMLR, 2018, pp. 639--648.

\bibitem{gama2004learning}
J.~Gama, P.~Medas, G.~Castillo, and P.~Rodrigues, ``Learning with drift
  detection,'' in \emph{Brazilian Symposium on Artificial Intelligence}.\hskip
  1em plus 0.5em minus 0.4em\relax Springer, 2004, pp. 286--295.

\bibitem{frias2014online}
I.~Frias-Blanco, J.~del Campo-{\'A}vila, G.~Ramos-Jimenez, R.~Morales-Bueno,
  A.~Ortiz-Diaz, and Y.~Caballero-Mota, ``Online and non-parametric drift
  detection methods based on {Hoeffding’s} bounds,'' \emph{IEEE Transactions
  on Knowledge and Data Engineering}, vol.~27, no.~3, pp. 810--823, 2014.

\bibitem{de2018wilcoxon}
R.~S.~M. de~Barros, J.~I.~G. Hidalgo, and D.~R. de~Lima~Cabral, ``Wilcoxon rank
  sum test drift detector,'' \emph{Neurocomputing}, vol. 275, pp. 1954--1963,
  2018.

\bibitem{ross2011nonparametric}
G.~J. Ross, D.~K. Tasoulis, and N.~M. Adams, ``Nonparametric monitoring of data
  streams for changes in location and scale,'' \emph{Technometrics}, vol.~53,
  no.~4, pp. 379--389, 2011.

\bibitem{gama2014survey}
J.~Gama, I.~{\v{Z}}liobait{\.e}, A.~Bifet, M.~Pechenizkiy, and A.~Bouchachia,
  ``A survey on concept drift adaptation,'' \emph{ACM Computing Surveys
  (CSUR)}, vol.~46, no.~4, p.~44, 2014.

\bibitem{wang2020auc}
S.~Wang and L.~L. Minku, ``{AUC} estimation and concept drift detection for
  imbalanced data streams with multiple classes,'' in \emph{2020 International
  Joint Conference on Neural Networks (IJCNN)}.\hskip 1em plus 0.5em minus
  0.4em\relax IEEE, 2020, pp. 1--8.

\bibitem{korycki2021concept}
{\L}.~Korycki and B.~Krawczyk, ``Concept drift detection from multi-class
  imbalanced data streams,'' in \emph{2021 IEEE 37th International Conference
  on Data Engineering (ICDE)}.\hskip 1em plus 0.5em minus 0.4em\relax IEEE,
  2021, pp. 1068--1079.

\bibitem{ross2012exponentially}
G.~J. Ross, N.~M. Adams, D.~K. Tasoulis, and D.~J. Hand, ``Exponentially
  weighted moving average charts for detecting concept drift,'' \emph{Pattern
  Recognition Letters}, vol.~33, no.~2, pp. 191--198, 2012.

\bibitem{roberts1959control}
S.~Roberts, ``Control chart tests based on geometric moving averages,''
  \emph{Technometrics}, vol.~1, no.~3, pp. 239--250, 1959.

\bibitem{guralnik1999event}
V.~Guralnik and J.~Srivastava, ``Event detection from time series data,'' in
  \emph{Proceedings of the 5$^{\text{th}}$ ACM SIGKDD International Conference
  on Knowledge Discovery and Data Mining}, 1999, pp. 33--42.

\bibitem{kuncheva2011change}
L.~I. Kuncheva, ``Change detection in streaming multivariate data using
  likelihood detectors,'' \emph{IEEE Transactions on Knowledge and Data
  Engineering}, vol.~25, no.~5, pp. 1175--1180, 2011.

\bibitem{alippi2015change}
C.~Alippi, G.~Boracchi, D.~Carrera, and M.~Roveri, ``Change detection in
  multivariate datastreams: Likelihood and detectability loss,''
  \emph{Proceedings of the International Joint Conference on Artificial
  Intelligence (IJCAI)}, vol.~2, pp. 1368--1374, 2016.

\bibitem{lau2018binning}
T.~S. Lau, W.~P. Tay, and V.~V. Veeravalli, ``A binning approach to quickest
  change detection with unknown post-change distribution,'' \emph{IEEE
  Transactions on Signal Processing}, vol.~67, no.~3, pp. 609--621, 2018.

\bibitem{lehmann2006testing}
E.~L. Lehmann and J.~P. Romano, \emph{Testing statistical hypotheses}.\hskip
  1em plus 0.5em minus 0.4em\relax Springer, 2006.

\bibitem{kuncheva2013pca}
L.~I. Kuncheva and W.~J. Faithfull, ``{PCA} feature extraction for change
  detection in multidimensional unlabeled data,'' \emph{IEEE Transactions on
  Neural Networks and Learning Systems}, vol.~25, no.~1, pp. 69--80, 2013.

\bibitem{qahtan2015pca}
A.~A. Qahtan, B.~Alharbi, S.~Wang, and X.~Zhang, ``A {PCA}-based change
  detection framework for multidimensional data streams,'' in \emph{Proceedings
  of the 21$^{\text{st}}$ ACM SIGKDD International Conference on Knowledge
  Discovery and Data Mining}, 2015, pp. 935--944.

\bibitem{ho2005martingale}
S.-S. Ho, ``A {Martingale} framework for concept change detection in
  time-varying data streams,'' in \emph{Proceedings of the 22$^{\text{nd}}$
  International Conference on Machine Learning}, 2005, pp. 321--327.

\bibitem{mozafari2011precise}
N.~Mozafari, S.~Hashemi, and A.~Hamzeh, ``A precise statistical approach for
  concept change detection in unlabeled data streams,'' \emph{Computers \&
  Mathematics with Applications}, vol.~62, no.~4, pp. 1655--1669, 2011.

\bibitem{gretton2012kernel}
A.~Gretton, K.~M. Borgwardt, M.~J. Rasch, B.~Sch{\"o}lkopf, and A.~Smola, ``A
  kernel two-sample test,'' \emph{The Journal of Machine Learning Research},
  vol.~13, no.~1, pp. 723--773, 2012.

\bibitem{li2015m}
S.~Li, Y.~Xie, H.~Dai, and L.~Song, ``M-statistic for kernel change-point
  detection,'' \emph{Advances in Neural Information Processing Systems},
  vol.~28, pp. 3366--3374, 2015.

\bibitem{keriven2020newma}
N.~Keriven, D.~Garreau, and I.~Poli, ``{NEWMA}: a new method for scalable
  model-free online change-point detection,'' \emph{IEEE Transactions on Signal
  Processing}, vol.~68, pp. 3515--3528, 2020.

\bibitem{margavio1995alarm}
T.~M. Margavio, M.~D. Conerly, W.~H. Woodall, and L.~G. Drake, ``Alarm rates
  for quality control charts,'' \emph{Statistics \& Probability Letters},
  vol.~24, no.~3, pp. 219--224, 1995.

\bibitem{souza2020challenges}
V.~Souza, D.~M. dos Reis, A.~G. Maletzke, and G.~E. Batista, ``Challenges in
  benchmarking stream learning algorithms with real-world data,'' \emph{Data
  Mining and Knowledge Discovery}, vol.~34, no.~6, pp. 1805--1858, 2020.

\bibitem{kullback1951information}
S.~Kullback and R.~A. Leibler, ``On information and sufficiency,'' \emph{The
  Annals of Mathematical Statistics}, vol.~22, no.~1, pp. 79--86, 1951.

\bibitem{demvsar2006statistical}
J.~Dem{\v{s}}ar, ``Statistical comparisons of classifiers over multiple data
  sets,'' \emph{The Journal of Machine Learning Research}, vol.~7, pp. 1--30,
  2006.

\bibitem{nemenyi1963distribution}
P.~B. Nemenyi, \emph{Distribution-free multiple comparisons}.\hskip 1em plus
  0.5em minus 0.4em\relax PhD Thesis, Princeton University, 1963.

\bibitem{dunn1961multiple}
O.~J. Dunn, ``Multiple comparisons among means,'' \emph{Journal of the American
  Statistical Association}, vol.~56, no. 293, pp. 52--64, 1961.

\end{thebibliography}
